%% file: root.tex
\documentclass[letterpaper, 10 pt, conference]{ieeeconf}  

\IEEEoverridecommandlockouts                              

\usepackage[caption=false,font=footnotesize]{subfig}
\usepackage{dblfloatfix}
\usepackage{graphics} 
\usepackage{graphicx}
\usepackage{epsfig} 
\usepackage{amsmath} 
\usepackage{amssymb}  
\usepackage{amsfonts}
\usepackage{color}
\usepackage{subfig}
\usepackage{adjustbox}
\usepackage{multirow}
\graphicspath{{Figures/}}
\usepackage{import}

\title{\LARGE \bf
Model Based Position Control of Soft Hydraulic Actuators
}

\author{Mark Runciman$^{1}$, Enrico Franco$^{2}$, James Avery$^{1}$,\\ 
Ferdinando Rodriguez y Baena$^{2}$, and George Mylonas$^{1}$ %
\thanks{This work was partially supported by a collaboration with the Multi-Scale Medical Robotics Centre, The Chinese University of Hong Kong. James Avery is an Imperial College Research Fellow. For the purpose of open access, the authors have applied a Creative Commons Attribution (CC BY) license to any Accepted Manuscript version arising.
}%
\thanks{$^{1}$The Hamlyn Centre, Imperial College London, London W2 1NY, UK. \textit{Corresponding author: Mark Runciman} {\tt\small m.runciman@imperial.ac.uk}}
\thanks{$^{2}$The Mechatronics in Medicine Laboratory, Mechanical Engineering Department, Imperial College London, London, SW7 2AZ UK.}
}

\begin{document}

\maketitle
\thispagestyle{empty}
\pagestyle{empty}

\begin{abstract}
In this article, we investigate the model based position control of soft hydraulic  actuators arranged in an antagonistic pair. A dynamical model of the system is constructed by employing the port-Hamiltonian formulation.
A control algorithm is designed with an energy shaping approach, which accounts for the pressure dynamics of the fluid.
A nonlinear observer is included to compensate the effect of unknown external forces.
Simulations demonstrate the effectiveness of the proposed approach, and experiments achieve positioning accuracy of 0.043 mm with a standard deviation of 0.033 mm in the presence of constant external forces up to 1 N.
\end{abstract}

\section{Introduction}\label{sec:1}
Soft robotic systems possess many of the features required in minimally invasive surgery (MIS), including low weight and compliance similar to that of biological systems \cite{Runciman2019}. In addition, soft robots allow for affordable designs by replacing expensive actuators with low-cost solutions that can be produced locally in a low-resource setting \cite{Franco2021a}.
Pneumatics and hydraulics are two common actuation strategies for soft robotic systems, due to their high power-to-weight ratio and affordability. In particular, pneumatic actuation yields fast responses and is well suited for force control, while hydraulic actuation enables the exertion of  higher forces. Both approaches have been used extensively, with pneumatics being the most common of the two \cite{Runciman2020}. Increasing attention has been focused on the design of soft hydraulic actuators, such as the one described in \cite{Runciman2021} for MIS applications, which combine high forces with a low-profile form factor, and the possibility to provide shape-sensing abilities based on the electrical impedance of the fluid \cite{Avery2019}.
Unlike pneumatic soft actuators \cite{Niiyama2015}, employing an incompressible fluid allows control of the length of the actuator in open-loop with good repeatability. Nevertheless, model based control becomes necessary if the application demands high position accuracy in the presence of unknown external forces.

Model-based control of soft robots is a notoriously complex topic, since these systems often possess more degrees-of-freedom (DOFs) than actuators \cite{Wang2022}. As a result, model-based control methods should account for the dynamics of the unactuated DOFs to ensure stability  \cite{Franco2021b,Borja2022}. In addition, the presence of disturbances, which are ubiquitous in unstructured environments, such as those commonly found in surgery, can degrade performance. To address this point, recent controllers for soft robots have included either nonlinear observers \cite{Franco2021b,Trumic2020a,Trumic2020b} or integral actions \cite{Franco2021c}.
Another challenge specific to soft robotic systems with pneumatic or hydraulic actuation is due to the pressure dynamics of the fluid, which decouples the control input from the dynamics of the payload \cite{Stolzle2021}. In our recent work \cite{Franco2022,Franco2022b,Franco2022c}, we have proposed an energy-based control approach for soft continuum manipulators that relies on the port-Hamiltonian formulation and extends the energy shaping methodology \cite{Ortega2002} by accounting for the internal energy of the fluid. Nevertheless, to the best of our knowledge, the case of multiple soft bellow actuators arranged in an antagonistic pair and subject to  disturbances has not yet been considered.

\begin{figure} [tb]
	\begin{center}
	\subfloat[\label{fig:1a}]{
		\begin{tabular}{c}
            {\tiny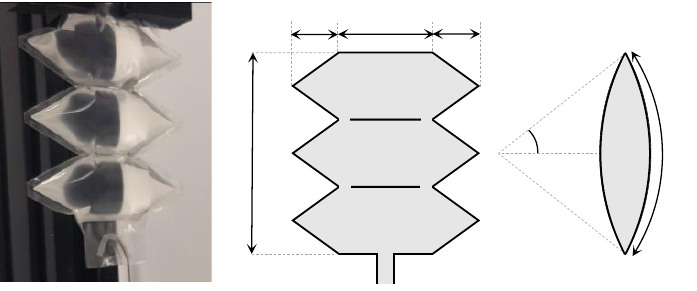}
		\end{tabular}
	}\\
	\subfloat[\label{fig:1b}]{
		\begin{tabular}{c}
            {\tiny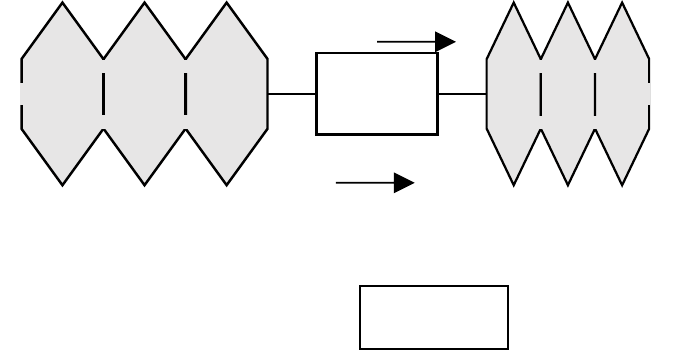}
		\end{tabular}
	}
		\caption{Depictions of soft hydraulic actuator (a); antagonistic pair (b).} \label{fig:fig1} \vspace{-0.7cm}
	\end{center}
\end{figure}

In this paper we investigate the model based control of soft hydraulic bellow actuators \cite{Runciman2021} arranged in an antagonistic pair (see Figure \ref{fig:fig1}) by employing a port-Hamiltonian formulation and an energy shaping control paradigm. 
The main contributions of this work include the following points.
\begin{itemize}
    \item A dynamical model of the soft hydraulic bellow actuator, which includes the pressure dynamics of the fluid, is presented. Differently from \cite{Runciman2021}, the relationship between the contraction of the actuator and its volume is expressed analytically in closed form and is employed for control purposes using an energy shaping procedure.
    \item A nonlinear observer is designed to compensate for the effect of unknown external forces in real-time. Stability conditions are discussed with a Lyapunov approach in relation to the tuning parameters. 
    \item The performance of the proposed controller is assessed with numerical simulations and extensive experiments.
\end{itemize}

The rest of the paper is organized as follows. Section \ref{sec:2} presents the system model. Section \ref{sec:3} details the controller design. Section \ref{sec:4} presents the results of simulations and experiments. 
Section \ref{sec:6} contains concluding remarks.

\section{System model}\label{sec:2}
We consider a system consisting of two soft hydraulic bellow actuators  \cite{Runciman2021}, denoted with the subscripts 1 and 2, supplied by pressurized water and arranged in an antagonistic pair to move a payload of mass \(m\) in the horizontal direction \(x\). The actuators are made of inextensible thermoplastic material (e.g. nylon) and, differently from pneumatic muscle actuators, they contract when the internal volume of the fluid increases.
Without loss of generality, we assume that the position \(x\) of the payload, which is connected to each actuator, increases when actuator 2 contracts and actuator 1 expands.
The actuators are supplied by identical syringe pumps, which are not modeled in detail this work.
The length of the bellow actuators varies with \(\theta\) as
\begin{equation} \label{eq:1}
    \begin{split}
        L(\theta)  = L_0 \frac{\sin{\theta}}{\theta},
    \end{split}
\end{equation}
where \(L_0\) is the length of the empty actuator, and \(\theta\) is half the central angle of the actuator's section when it is contracted (see Figure \ref{fig:1a}).
The volume of one bellow actuator varies in a nonlinear fashion with \(\theta\), that is
\begin{equation} \label{eq:2}
    \begin{split}
        V(\theta)  = k_0 \frac{L_0^2}{n_L}\left(\frac{d_c}{3}+\frac{D_s}{2}\right) \frac{\theta- \cos{\theta}\sin{\theta}}{\theta^2},
    \end{split}
\end{equation}
where \(n_L\) is the number of pouches in the actuator, \(d_c\) and \(D_s\) define the actuator's geometry, and \(k_0\) is a scaling factor \cite{Runciman2021}.
A closed-form analytical expression that approximates the volumes \(V_1\) and \(V_2\) of the antagonistic pair (see Figure \ref{fig:1b}) is obtained by substituting Taylor series in (\ref{eq:1}) and (\ref{eq:2}), that is \(\sin{\theta}\approx \theta - \frac{\theta^3}{6}\) and \(\cos{\theta}\approx 1 - \frac{\theta^2}{2}\), which yields \(L(\theta)  = L_0 (1-\frac{\theta^2}{6})\). Defining the contraction of the actuator \(V_2\)  as \(x=L_0-L(\theta)\), the volumes \(V_1\) and \(V_2\) yield
\begin{equation} \label{eq:3}
    \begin{split}
        V_1  = K_0 \left(\frac{2}{3}-\frac{x_M-x-x_0}{2L_0}\right)\sqrt{\frac{6(x_M-x-x_0)}{L_0}}+V_0,\\
        V_2  = K_0 \left(\frac{2}{3}-\frac{x+x_0}{2L_0}\right)\sqrt{\frac{6(x+x_0)}{L_0}}+V_0,
    \end{split}
\end{equation}
where \(K_0\) is a scaling factor that accounts for the parameters in (\ref{eq:2}), \(x_0\) is the initial position, \(x_M\) is the maximum contraction of the actuators, and \(V_0\) is the dead volume of fluid assumed to be identical for both actuators.

The mechanical energy \(H\) of the system includes the kinetic energy of the payload and of the fluid, and the internal energy of the pressurized fluid \(\Phi\) in each actuator. The potential elastic energy is instead negligible, since the actuator material does not stretch longitudinally and the system lies on the horizontal plane. In summary, \(H = \frac{1}{2}M\dot{x}^2 + \Phi_1 + \Phi_2\), where the internal energy of the pressurized fluid in each bellow actuator is \cite{Gao2019}
\begin{equation} \label{eq:4}
\begin{split}
\Phi_1 = \left(- P_1 + \Gamma_{0}(e^{P_1/\Gamma_{0}}-1)\right)V_1,\\
\Phi_2 = \left( -P_2 + \Gamma_{0}(e^{P_2/\Gamma_{0}}-1)\right)V_2,
\end{split}
\end{equation}
and the pressures \(P_1\) and \(P_2\) are relative to atmosphere, while the total mass of the moving parts for a fluid of constant density \(\rho\) is
\begin{equation} \label{eq:5}
\begin{split}
M  = \left( m + V_1 \rho + V_2 \rho\right).
\end{split}
\end{equation}
Denoting the isothermal bulk modulus of the fluid with \(\Gamma_0\), the pressure dynamics are given by
\begin{equation} \label{eq:6}
\begin{split}
\dot{P}_1 = \Gamma_{0} \frac{U_1 - A_1 \dot{x}}{ V_{1}}, ~
\dot{P}_2 = \Gamma_{0} \frac{U_2 - A_2 \dot{x}}{ V_{2}},
\end{split}
\end{equation}
where the volumetric flow rates \(U_1\) and \(U_2\) provided by the syringe pumps correspond to the control input \cite{Acuna-Bravo2009}, while \(A_1=\frac{\partial V_1}{\partial x}\) and \(A_2=\frac{\partial V_2}{\partial x}\).
The system dynamics in port-Hamiltonian form, without the internal dynamics of the syringe pumps, is thus
\begin{equation} \label{eq:7}
\begin{split}
\begin{bmatrix}
\dot{x} \\
\dot{p} \\
\dot{P_1} \\
\dot{P_2} \\
\end{bmatrix} = \begin{bmatrix}
0 & 1 & 0 & 0\\
 - 1 & - R & \Gamma_{01} & \Gamma_{02}\\
0 & -\Gamma_{01} & 0 & 0 \\
0 & -\Gamma_{02} & 0 & 0 \\
\end{bmatrix}\begin{bmatrix}
\partial_x H \\
\partial_p H \\
\partial_{P_1} H \\
\partial_{P_2} H \\
\end{bmatrix} 
+ \begin{bmatrix}
0 \\
-F \\
\frac{\Gamma_{0}U_1}{V_1}\\
\frac{\Gamma_{0}U_2}{V_2}\
\end{bmatrix}, 
\end{split}
\end{equation}
where \(\Gamma_{01} = \frac{\Gamma_{0}A_1}{V_1} \) and \(\Gamma_{02} = \frac{\Gamma_{0}A_2}{V_2} \), \(R\) is the physical damping related to the transmission, and \(F\) is the external force due to the payload.
The system states are the position \(x\) of the payload, the momenta \(p = M\dot{x}\), and the pressures \(P_1\) and \(P_2\). The notation \(\partial_x H=\frac{\partial H}{\partial x}, \partial_p H=\frac{\partial H}{\partial p}, \partial_{P_1} H=\frac{\partial H}{\partial P_1}, \partial_{P_2} H=\frac{\partial H}{\partial P_2}\) is employed for brevity.
The following assumptions are introduced for controller design purposes.

\emph{Assumption 1}. The fluid is isothermal, isentropic, and inviscid. The pressures \(P_1\) and \(P_2\), the density \(\rho\) (assumed constant), and the speed of the fluid (which is approximated with \(\dot{x}\)) are uniform throughout the volumes \(V_1\) and \(V_2\).

\emph{Assumption 2}. All model parameters are accurately known. The bulk modulus of the fluid is \(\Gamma_0\). The friction of the transmission (i.e. the lead-screw of the syringe pump, and the cable attached to the payload) is defined by the parameter \(R\). The system lies in the horizontal plane.

\emph{Assumption 3}. The position \(x\) and the velocity \(\dot{x}\) of the payload, and the pressures \(P_1\) and \(P_2\) of the fluid are measurable and bounded, that is \(P_1,P_2 \ll \Gamma_0 \) .

\emph{Assumption 4}. The effect of the external forces is accounted for with \(F\), which is unknown but constant and can be either positive or negative.

The effect of viscosity on the pressure dynamics is negligible at low speed \cite{Franco2022}, while pressure, speed and density are near-uniform in the case of laminar flow. 
Constant external forces can include the weight of an additional payload, while the case of time-varying forces is discussed in Section \ref{sec:3.3}.

\section{Controller design}\label{sec:3}
The control goal corresponds to regulating the position of the payload to \(x=x^*\) in the presence of an unknown external force \(F\).

\subsection{Nonlinear observer}\label{sec:3.1}
The external force is estimated with a nonlinear observer constructed according to the \emph{Immersion and Invariance} methodology \cite{AstolfiA.KaragiannisD.2007}.
To this end, the estimation error \(\zeta\) is defined as
\begin{equation} \label{eq:8}
\begin{split}
\zeta = {\widehat{F}} + \beta - F,
\end{split}
\end{equation}
where the force estimate is \(\Tilde{F} =\widehat{F} + \beta \). The function \(\beta\), which is the state-dependent part of the force estimate, and the observer state \(\widehat{F}\) are computed with
\begin{equation} \label{eq:9}
\begin{split}
{\dot{\widehat{F}}} = \alpha \left(-\partial_x H -R \partial_p H + \frac{\Gamma_{0}A_1}{V_1} \partial_{P_1} H \right) \\
+\alpha \left( \frac{\Gamma_{0}A_2}{V_2} \partial_{P_2} H  - {\widehat{F}} - \beta \right)
,\\
\beta = -\alpha p,
\end{split}
\end{equation}
with \(\alpha>0\) a constant tuning parameter.

\emph{Proposition 1}: Consider system (\ref{eq:7}) with \emph{Assumptions 1} to
\emph{4} and with the observer (\ref{eq:9}). Then \(\zeta\) converges to zero exponentially for all \(\alpha >0\).

\emph{Proof}: Computing the time derivative of (\ref{eq:8}) while substituting \(\dot{p}, \dot{P_1}\) and \(\dot{P_2}\) from (\ref{eq:7}) gives
\begin{equation} \label{eq:10}
\begin{split}
\dot{\zeta}={\dot{\widehat{F}}} + \frac{\partial \beta}{\partial x} \frac{p}{M} + \frac{\partial \beta}{\partial p} 
\left(-\partial_x H - {\widehat{F}} - \beta + \zeta\right)\\
+ \frac{\partial \beta}{\partial p}  \left( \frac{\Gamma_{0}A_1}{V_1} \partial_{P_1} H + \frac{\Gamma_{0}A_2}{V_2} \partial_{P_2} H  -R \partial_p H \right). 
\end{split}
\end{equation}
Substituting (\ref{eq:9}) into (\ref{eq:10}) yields
\begin{equation} \label{eq:11}
\begin{split}
\dot{\zeta} = -\alpha \zeta.\
\end{split}
\end{equation}
Defining the Lyapunov function candidate \(\Upsilon = \frac{1}{2}\zeta^{2}\), computing its time derivative, and substituting (\ref{eq:11}) yields
\begin{equation} \label{eq:12}
\begin{split}
\dot{\Upsilon} = - \alpha \zeta^{2}  = -2 \alpha \Upsilon <0.
\end{split}
\end{equation}
It follows from (\ref{eq:12}) that \(\zeta\) is bounded and converges to zero exponentially for all \( \alpha > 0\) concluding the proof \(\square\)

\subsection{Energy shaping control}\label{sec:3.2}
The control law is designed following a similar procedure to \cite{Franco2022}, which is extended to account for the presence of redundant actuators in the antagonistic pair and for the nonlinear observer  (\ref{eq:9}). The closed-loop dynamics in port-Hamiltonian form yields thus
\begin{equation} \label{eq:13}
\begin{bmatrix}
\dot{x} \\
\dot{p} \\
\dot{P_1} \\
\dot{P_2} \\
\end{bmatrix} = \begin{bmatrix}
 0 & S_{12} & S_{13} & S_{14}\\
 - S_{12} & - S_{22} & S_{23} & S_{24}\\
 - S_{13} & - S_{23} & - S_{33} & 0\\
 - S_{14} & - S_{24} & 0 & -S_{44}\\
\end{bmatrix}\begin{bmatrix}
\partial_x H_d \\
\partial_p H_d \\
\partial_{P_1} H_d \\
\partial_{P_2} H_d \\
\end{bmatrix} - \begin{bmatrix}
0 \\
\zeta \\
0 \\
0 \\
\end{bmatrix},\
\end{equation}
where \(H_{d} = \frac{1}{2} p^{2}M_{d}^{- 1}  + \Omega_{d} + \varsigma^{2}/2\) is a positive definite storage function. The potential energy \(\Omega_{d} = \frac{1}{2}k_p \left(x^* -x \right)^2\) has a strict minimizer at \(x=x^*\) corresponding to the regulation goal, \(M_{d} = k_{m}M\), and \(\varsigma\) is given by
\begin{equation} \label{eq:14}
\begin{split}
\varsigma =P_1 A_1 + P_2 A_2 - {\widehat{F}} + k_p k_m\left(x-x^* \right),
\end{split}
\end{equation}
with \(k_{p}>0\) and \(k_m>0\) constant tuning parameters, \(A_1=\frac{\partial V_1}{\partial x}\) and \(A_2=\frac{\partial V_2}{\partial x}\). and \({\widehat{F}}\) is computed by time-integration of (\ref{eq:9}). The terms \(S_{\text{ij}}\) are defined so that the open-loop dynamics (\ref{eq:7}) matches the closed-loop dynamics (\ref{eq:13}) accounting for the estimation error (\ref{eq:8}), that is
\begin{equation} \label{eq:15}
\begin{split}
S_{12} = k_{m}, ~S_{13}=S_{14}=0, ~S_{22} = k_{m}R - \alpha k_m M, \\
S_{23} = \frac{ 1 + k_{m}\partial_{x}\varsigma }{2\partial_{P_1}\varsigma}, ~
S_{24} = \frac{ 1 + k_{m}\partial_{x}\varsigma }{2\partial_{P_2}\varsigma}, \\
S_{33} = \frac{k_{i}}{\left(\partial_{P_1}\varsigma\right)^2}>0, ~
S_{44} = \frac{k_{i}}{\left(\partial_{P_2}\varsigma\right)^2} > 0.
\end{split}
\end{equation}
The control inputs \(U_1\) and \(U_2\) are thus
\begin{equation} \label{eq:16}
\begin{split}
U_1 = \frac{A_1 p}{M} - \frac{V_{1} }{\Gamma_{0}} \left( \frac{ 1 + k_{m}\partial_{x}\varsigma }{2 A_1} \frac{p}{M} +  \frac{k_{i}}{A_1}\varsigma \right), \\
U_2 = \frac{A_2 p}{M} - \frac{ V_{2}}{\Gamma_{0}} \left( \frac{ 1 + k_{m}\partial_{x}\varsigma }{2 A_2} \frac{p}{M} +  \frac{k_{i}}{A_2} \varsigma \right),
\end{split}
\end{equation}
where the tuning parameters are \(k_m, k_p, k_i, \text{and}~ \alpha\) in (\ref{eq:9}).

\emph{Lemma 1}: The system (\ref{eq:7}) in closed-loop with the control laws (\ref{eq:16}) yields (\ref{eq:13}) with the parameters (\ref{eq:14}) and (\ref{eq:15}). 

\emph{Proof}: Equating the corresponding rows of (\ref{eq:7}) and of (\ref{eq:13}) yields the matching equations
\begin{equation} \label{eq:17}
\begin{split}
M^{- 1}p = S_{12}M_{d}^{- 1}p + S_{13} \varsigma \partial_{P_1} \varsigma + S_{14} \varsigma \partial_{P_2} \varsigma, \\
\frac{\Gamma_{0}A_1\partial_{P_1} H}{V_1}  + \frac{\Gamma_{0}A_2\partial_{P_2} H}{V_2} -\partial_x H -R \partial_p H -F = \\
- {S}_{12}\left( \partial_{x}\Omega_{d} + \frac{1}{2}\partial_{x}( p^{T}M_{d}^{- 1}p) + \varsigma \partial_x\varsigma\right) \\
- S_{22}M_{d}^{- 1}p + S_{23} \varsigma \partial_{P_1}\varsigma+ S_{24}\varsigma \partial_{P_2} \varsigma - \zeta - \alpha p, \\
\frac{ \Gamma_{0} }{V_{1}} \left(U_1-A_1 \frac{p}{M}\right) = -S_{13} \varsigma \partial_{x} \varsigma - S_{23} \frac{p}{M_d} - S_{33}\varsigma \partial_{P_1}\varsigma, \\
\frac{ \Gamma_{0} }{ V_{2}} \left(U_2-A_2 \frac{p}{M}\right)= -S_{14} \varsigma \partial_{x} \varsigma - S_{24} \frac{p}{M_d}  - S_{44}\varsigma \partial_{P_2}\varsigma,
\end{split}
\end{equation}
which are verified by the parameters (\ref{eq:15}).
In particular, the first equation is verified by \(S_{12} = k_{m}\) and \(M_{d} = k_{m}M\) with \(S_{13} = S_{14} = 0\) since \(\partial_p\varsigma=0\). 
Substituting \(S_{12}\), \(S_{22}\), \(S_{23}\) and \(S_{24}\) verifies the second equation with \(\varsigma\) in (\ref{eq:14}).
Finally, substituting \(U_1\) and \(U_2\) from (\ref{eq:16}) with \(S_{13}=S_{14}=0\) and \(\partial_{P_1}\varsigma=A_1, \partial_{P_2}\varsigma=A_2\) verifies the last two equations \(\square\)

\emph{Remark 1}. Differently from our previous work \cite{Franco2021b,Franco2022}, the system (\ref{eq:7}) is fully actuated. Thus, the controller design does not require solving partial differential equations, which is a major challenge in energy shaping control \cite{Ortega2002}. However, the payload dynamics are not input-affine due to the pressure dynamics of the fluid, which is similar to our work \cite{Franco2022}. 
In this regard, the first key difference from \cite{Franco2022} is due to the presence of redundant bellow actuators in the antagonistic pair that are characterized by nonlinear expressions of the volumes \(V_1\) and \(V_2\), which yields nonlinear control laws.
The second key difference is due to the nonlinear observer (\ref{eq:9}) which results in the closed-loop damping \(S_{22}\) in (\ref{eq:15})  including a negative term proportional to \(\alpha\). This so-called negative damping assignment greatly simplifies the controller design. 
For comparison purposes, redefining (\ref{eq:14}) as
\begin{equation} \notag
\begin{split}
\varsigma =P_1 A_1 + P_2 A_2 - {\widehat{F}} + \alpha p + k_p k_m\left(x-x^* \right),
\end{split}
\end{equation}
would cancel the term \(\alpha p\) from the second equation in (\ref{eq:17}) yielding \(S_{22}=k_m R\)  as in \cite{Franco2022}.
This would lead to a more complex control law, since \(\partial_p \varsigma \neq 0\) thus requiring \(S_{13}\neq 0\) to verify the first matching equation in (\ref{eq:17}), that is
\begin{equation} \notag
\begin{split}
M^{- 1}p = S_{12}M_{d}^{- 1}p + S_{12} \varsigma \partial_p \varsigma + S_{13} \varsigma \partial_{P_1} \varsigma + S_{14} \varsigma \partial_{P_2} \varsigma.
\end{split}
\end{equation}

\emph{Remark 2}. The dynamics of the syringe pumps supplying the flow rates \(U_1\) and \(U_2\) is not modeled for simplicity. However, in case the syringe pumps are actuated by stepper motors, the control laws (\ref{eq:16}) can be employed to design a reference trajectory \(x_s (t)\). For instance, employing a minimum-jerk trajectory with duration \(T_f\) yields
\begin{equation} \notag
\begin{split}
x_s (t) = x_{s0} + (x_s^* - x_{s0})\left(\frac{10 t^3}{T_f^3}-\frac{15 t^4}{T_f^4}+\frac{6 t^6}{T_f^6}\right),
\end{split}
\end{equation}
where \(x_{s0}\) and \(x_s^*\) are the initial position and the final position of the stepper motor.  
Computing the time derivative of \(x_s (t)\), while noting that the flow rate of a syringe pump with area \(S\) is \(U_1=\dot{x}_s S\) and corresponds to the control input (\ref{eq:16}), yields the target position for the first stepper motor at any instant \(0<t<T_f\) 
\begin{equation} \notag
\begin{split}
x_s^* = x_{s0} + \frac{U_1 T_f^5 }{30 t^2 S (T_f - t)^2}.
\end{split}
\end{equation}
In a digital implementation with sampling interval \(\Delta t\), the former expression is modified by substituting \(t=\Delta t\) (i.e. \(x_s^*\) is computed at each instant only for the subsequent sampling interval).
If in addition \(T_f = 2\Delta t\) we have
\begin{equation} \notag
\begin{split}
x_s^* = x_{s0} + \frac{32 U_1 \Delta t  }{30 S},
\end{split}
\end{equation}
where \(U_1\) is given by (\ref{eq:16}). 

\subsection{Stability analysis}\label{sec:3.3}
\emph{Proposition 2}: Consider system (\ref{eq:7}) with \emph{Assumptions 1} to \emph{4} in closed-loop with the control laws (\ref{eq:16}), where the adaptive estimate of the force \({\Tilde{F}}=\widehat{F} - \alpha p\) is computed with (\ref{eq:9}).
Define the parameters
\(k_{i},k_{m},\alpha\) such that the matrix
\begin{equation} \label{eq:18}
\begin{split}
\Theta = 
\begin{bmatrix}
\frac{R - \alpha M}{k_m M^2} & \frac{1}{2 k_m M} & 0  \\
\frac{1}{2 k_m M} & \alpha & 0  \\
0 & 0 & 2k_{i}\\
\end{bmatrix},\
\end{split}
\end{equation}
is positive definite, that is \(k_i>0, (R - \alpha M)\alpha k_m > \frac{1}{4} \). Then the equilibrium point
\((x,\dot{x},P_1,P_2) = \left(x^{*},0,P_1^*,P_2^*\right)\) is globally asymptotically stable provided that \(A_1 \neq -A_2\).

\emph{Proof}: Defining the Lyapunov function \(\Psi = {H}_{d} + \Upsilon\) and computing its time derivative along the trajectories of the closed-loop system (\ref{eq:13}) while substituting (\ref{eq:12}) yields
\begin{equation} \label{eq:19}
\begin{split}
\dot{\Psi} = - S_{22}\left(\partial_{p}H_{d}\right)^2 -\partial_{p}H_{d}\zeta - \alpha \zeta^{2}
- 2 k_i\varsigma^{2}.
\end{split}
\end{equation}
Refactoring common terms in (\ref{eq:19}) yields
\begin{equation} \label{eq:20}
\begin{split}
\dot{\Psi} = - \overline{x}^{T}\Theta \overline{x},\
\end{split}
\end{equation}
where \(\overline{x}^{T} = \begin{bmatrix}
p & \zeta & \varsigma \end{bmatrix}\) and \(\Theta\) is given in (\ref{eq:18}). Thus \(\dot{\Psi} \leq 0\) for all \(k_i>0, (R - \alpha M) \alpha k_m > \frac{1}{4} \) and the equilibrium is stable. 
It follows from (\ref{eq:20}) that \(\overline{x} \in \mathcal{L}^{2} \cap \mathcal{L}^{\infty}\), while computing \(\dot{p}\) from (\ref{eq:13}) yields \(\dot{p} \in \mathcal{L}^{\infty} \). Similarly, it follows from (\ref{eq:14}) that \(\dot{\varsigma} \in \mathcal{L}^{\infty} \), and \(\dot{\zeta} \in \mathcal{L}^{\infty} \) from (\ref{eq:11}). Consequently, \(\overline{x}^{T}\) converges to zero asymptotically \cite{Tao1997}. 
Computing \(\dot{p}\) from (\ref{eq:13}) at \(\overline{x} = 0\) yields \(\partial_{x}\Omega_{d} = 0\), that is \( k_p k_m (x^*-x)=0\). In addition, \(\partial_{x}^2\Omega_{d} = k_p k_m>0\) which confirms that the equilibrium is a strict minimizer of \(\Omega_{d}\).  
Thus \(x=x^*\) is the largest invariant set in \(\overline{x}=0\) and it is asymptotically stable (see Corollary 3.1 in \cite{Khalil2002}).
Finally, computing (\ref{eq:14}) at \(\overline{x} = 0\) and \(x=x^*\) yields the values of \(P_1^*,P_2^*\)
at the equilibrium, that is \(P_1^* A_1 + P_2^* A_2 = {\widehat{F}}\).

To prove the global claim it is necessary to show that \(\Psi\) is radially unbounded (see Corollary 3.2 in \cite{Khalil2002}).
To this end, note that \(x,p,\varsigma,\zeta \rightarrow \infty \implies \Psi \rightarrow \infty\). In addition, it follows from (\ref{eq:14}) that \(P_1,P_2 \rightarrow \infty \implies \varsigma \rightarrow \infty\) provided that \(A_1 \neq -A_2\), while \(\varsigma=0 \implies A_1 P_1 + A_2 P_2 = k_p(x^* -x)+ \widehat{F}\). 
Consequently, provided that \(A_1 \neq -A_2\), the condition \(\varsigma=0 \cap P_1, P_2 \rightarrow \infty\) requires either \(x \rightarrow \infty\) or \(\widehat{F} \rightarrow \infty\), which yield \(\Psi \rightarrow \infty\) and concludes the proof \(\square\)

\emph{Remark 3}. The negative damping assignment in \(S_{22}\) imposes an upper bound on \(\alpha\), that is \(0<\alpha<R/M\).
If the force \(F\) is time-varying with time derivative \(\dot{F} = c \dot{x}\), where \(0 \leq c \leq \epsilon\), equation (\ref{eq:19}) yields 
\begin{equation} \notag
\begin{split}
\dot{\Psi} \leq - S_{22}\left(\partial_{p}H_{d}\right)^2 -\partial_{p}H_{d}\zeta - \alpha \zeta^{2} - \zeta \epsilon \frac{p}{M} - 2 k_i\varsigma^{2},
\end{split}
\end{equation}
which can be written as (\ref{eq:20}) with the new matrix
\begin{equation} \label{eq:21}
\begin{split}
\Theta^{'} = 
\begin{bmatrix}
\frac{R - \alpha M}{k_m M^2} & \frac{1}{2 k_m M} + \frac{\epsilon}{2 M} & 0  \\
\frac{1}{2 k_m M} + \frac{\epsilon}{2 M} & \alpha & 0  \\
0 & 0 & 2k_{i}\\
\end{bmatrix}.\
\end{split}
\end{equation}
Global asymptotic stability of the equilibrium is concluded if \(k_i>0\) and \( (R- \alpha M) \alpha k_m > \frac{1}{4} \left(1+ \epsilon k_m\right)^2\). 
Rearranging the former inequality yields
\(\epsilon^2 k_m^2 -  2 k_m (-\epsilon + 2 (R - \alpha M)\alpha) +1 < 0 \), which admits real solutions provided that \((R - \alpha M)\alpha > \epsilon/2 \): this indicates that the presence of time varying external forces requires either a larger physical damping \(R\) or a less aggressive tuning of the parameter \(\alpha\).

\begin{figure} [ht]
	\begin{center}
	\subfloat[\label{fig:2a}]{
		\begin{tabular}{c}
			\includegraphics[scale = 0.6]{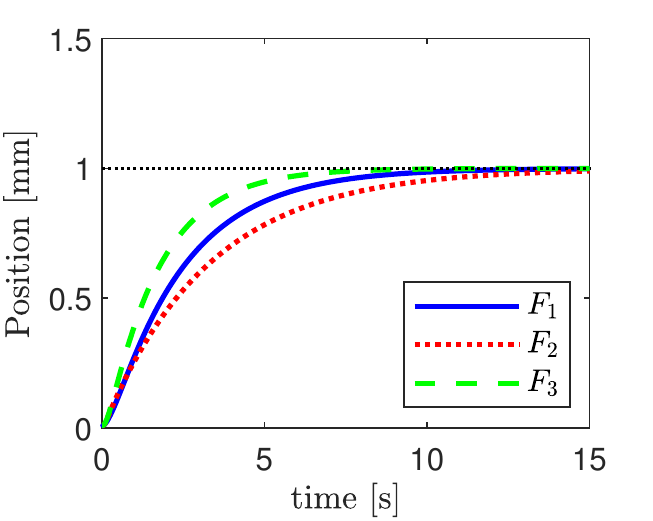} 
		\end{tabular}
	}
	\hspace*{-2.7em}
	\subfloat[\label{fig:2b}]{
		\begin{tabular}{c}
			\includegraphics[scale = 0.6]{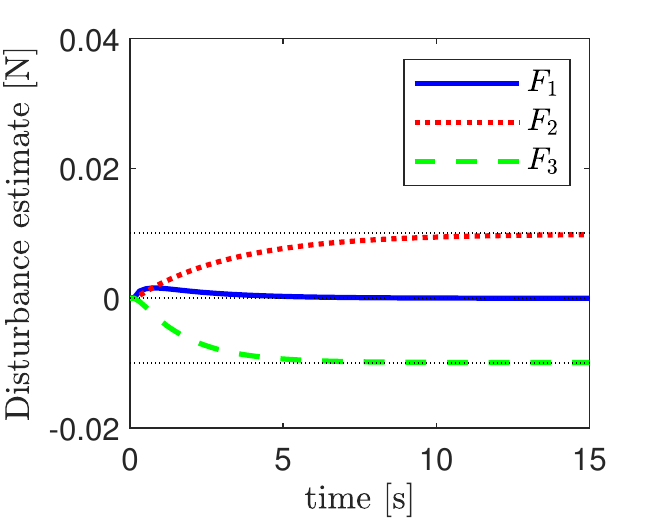} 
		\end{tabular}
	}\\[-0.3ex]
	\subfloat[\label{fig:2c}]{
		\begin{tabular}{c}
			\includegraphics[scale = 0.6]{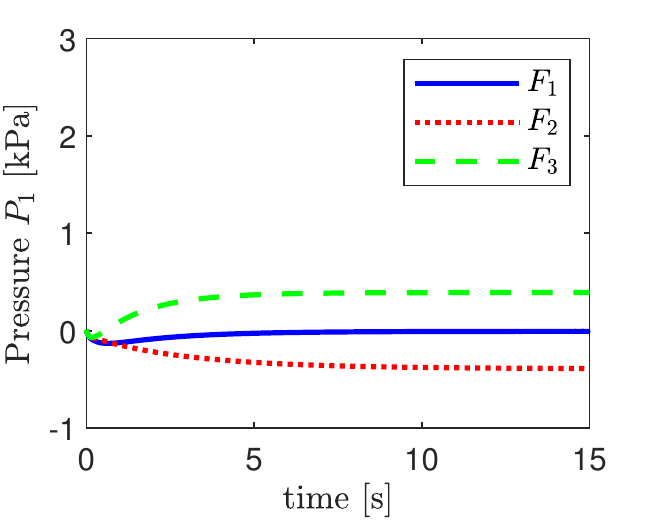} 
		\end{tabular}
	}
	\hspace*{-2.7em}
	\subfloat[\label{fig:2d}]{
		\begin{tabular}{c}
			\includegraphics[scale = 0.6]{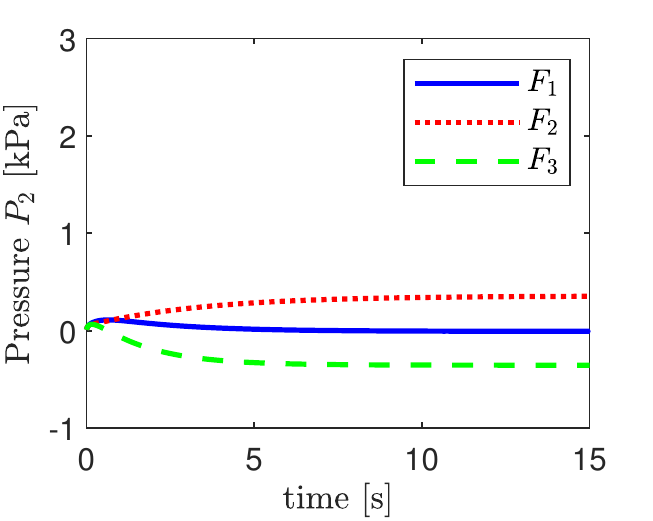} 
		\end{tabular}
	}\\[-0.3ex]
	\subfloat[\label{fig:2e}]{
		\begin{tabular}{c}
			\includegraphics[scale = 0.6]{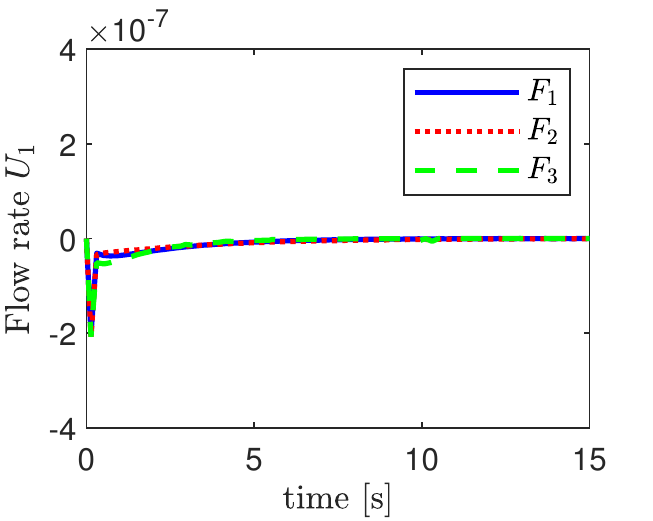} 
		\end{tabular}
	}
	\hspace*{-2.7em}
	\subfloat[\label{fig:2f}]{
		\begin{tabular}{c}
			\includegraphics[scale = 0.6]{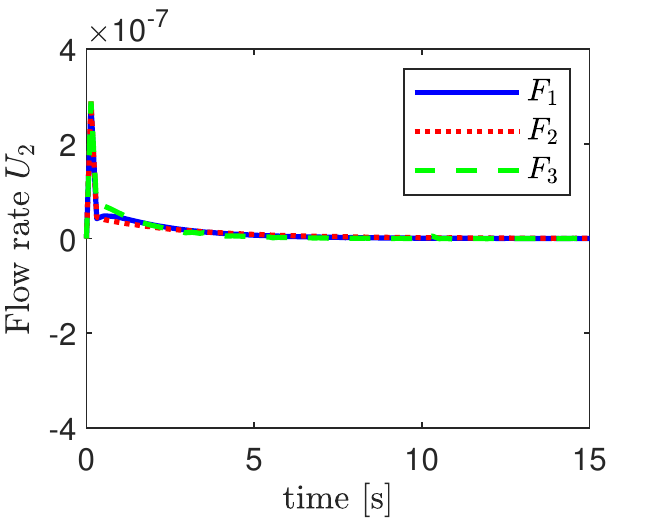} 
		\end{tabular}
	}
	~
		\caption{Simulation results for system (\ref{eq:7}) and controller (\ref{eq:16}) considering different external forces, \(F_1=5 \tanh{(\dot{x})}\), \(F_2=10 x\), \(F_3=-10 x\) : (a) position \(x\); (b) disturbance estimate \(\Tilde{F}\); (c) pressure \(P_1\); (d) pressure \(P_2\); (e) control input \(U_1\); (f) control input \(U_2\).} \label{fig:2} \vspace{-0.5cm}
	\end{center}
\end{figure}

\section{Results}\label{sec:4}

\subsection{Simulations}\label{sec:4.1}
Simulations have been conducted in MATLAB using an ODE23 solver and the initial conditions \((x,\dot{x},P_1,P_2)=(0,0,0,0)\) with atmospheric pressure \(P_{\text{atm}}=10^5\). 
The model parameters in SI units are \(\Gamma_0=2 \times 10^9, ~\rho=10^3, ~R=5, ~m=0.25, ~L_0=30 \times 10^{-3}, ~V_0=1\times 10^{-7}, ~n_L=3, ~D_s=12\times 10^{-3}, ~d_c=9\times 10^{-3}, ~x_0=L_0/8\), \(x_M=L_0/4\), and \(K_0=1.05 \frac{L_0^2}{n_L} (\frac{d_c}{3}+\frac{D_s}{2})=2.8 \times 10^{-6}\).
The tuning parameters for the control laws (\ref{eq:16}) have been set as \( k_p=1, k_m=2, k_i=10, \alpha=10\) for illustrative purposes. Note that the former values verify the stability conditions of \emph{Proposition 2}, that is \(k_i=10>0, (R -\alpha M)\alpha k_m = 80 > \frac{1}{4}\).
Three different external forces have been considered: \(F=5 \tanh{(\dot{x})}\), which is akin to Coulomb friction and vanishes at equilibrium, is indicated with \(F_1\); \(F=10 x\), which represents a compression spring, is indicated with \(F_2\); \(F=-10 x\), which represents a tension spring, is indicated with \(F_3\).

Figure \ref{fig:2} shows that the regulation goal \(x=x^*\) is correctly achieved with the control laws (\ref{eq:16}) in spite of the different external forces. In particular, the forces \(F_1\) and \(F_2\) oppose motion, while \(F_3\) favours motion, thus resulting in a slightly faster transient. The disturbance observer (\ref{eq:9}) converges to a constant value, which corresponds to the forces \(F_1\), \(F_2\), and \(F_3\) at equilibrium, that is \(0\) N, \(0.01\) N, \(-0.01\) N.
The control inputs and the corresponding pressures remain smooth for all operating conditions.

\subsection{Experiments}\label{sec:4.2}
\begin{figure}[tb]
	\def\svgwidth{\columnwidth}
	\centering
    {\footnotesize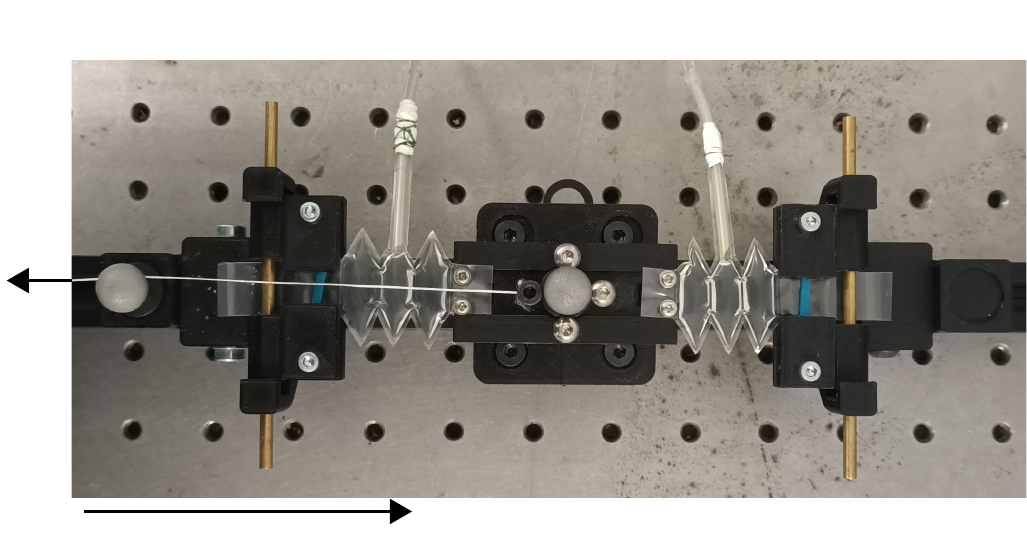}
    \caption{Experimental setup, showing antagonistic arrangement of soft actuators, the gantry plate, optical trackers, and cable used to transmit external force $F_{ext}$ to gantry.}\vspace{-0.5cm}
    \label{fig:3}
\end{figure}

The controller (\ref{eq:16}) was tested experimentally on a prototype consisting of two identical soft hydraulic actuators arranged in an antagonistic pair (see Figure \ref{fig:3}). The actuator dimensions are the same as those specified in Section \ref{sec:4.1}. 
The actuators are supplied by two identical syringe pumps (ID 27 mm) driven by stepper motors and lead-screw transmission. The position \(x\)  has been measured with an optical tracking system (OptiTrack, NaturalPoint, Inc., USA), and the pressures have been measured with two sensors (MS5803-14BA, TE Connectivity, Switzerland).
A Python script was employed to collect data from the sensors via serial link with baud rate 115200 and to communicate with the stepper drivers (DRV8825, Pololu, USA) with a sampling frequency of 20.84 Hz.
The position command issued to the stepper \(j=1,2\) has been computed from (\ref{eq:16}) as in \emph{Remark 2}, that is \(x_{sj}^*=x_{sj0} + U_{j} k_U\), where \(k_U\) is a constant depending on the size of the syringe pumps. The tuning parameters have been set to \( k_p=4, k_m=4, k_i=10, \alpha=10\), which are similar to those used in the simulations, while \(k_U=6\) has been chosen empirically.
The starting position corresponding to \(x=0\) has been set empirically by filling both actuators by equal amounts so that \(x_0\approx L_0/8\), after which the controller has been activated.
To assess the effect of external forces, different masses (i.e., 50 g, 75 g, 100 g) have been attached to the gantry plate with a pulley, thus resulting in a constant force in the negative $x$ direction.

\begin{figure} [tb]
	\begin{center}
	\subfloat[\label{fig:4a}]{
		\begin{tabular}{c}
			\includegraphics[scale = 0.6]{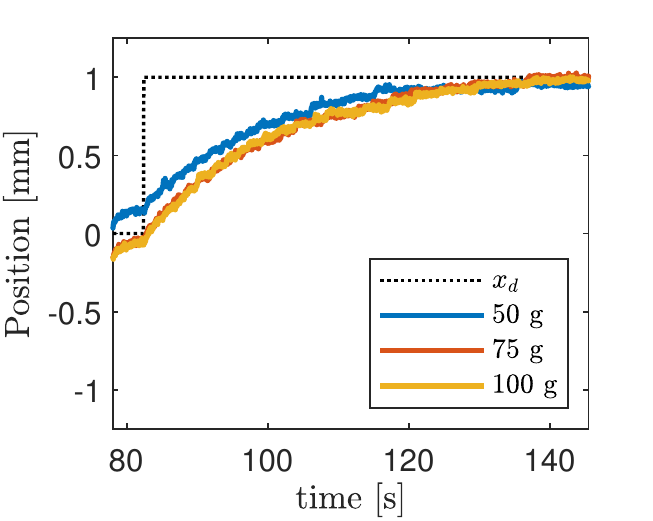} 
		\end{tabular}
	}
	\hspace*{-2.7em}
	\subfloat[\label{fig:4b}]{
		\begin{tabular}{c}
			\includegraphics[scale = 0.6]{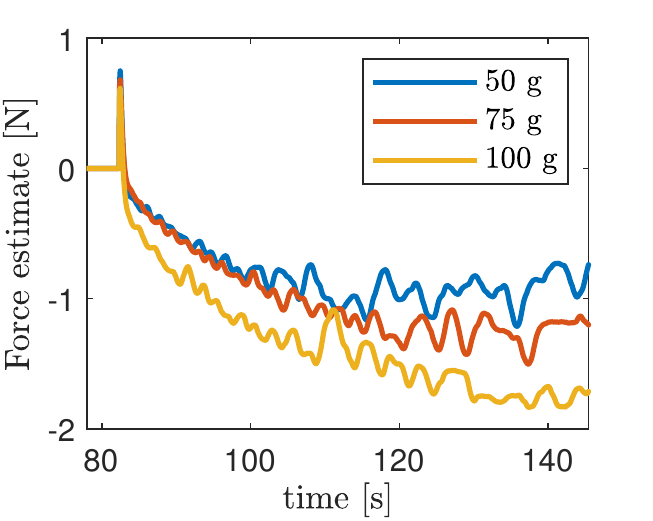} 
		\end{tabular}
	}\\[-0.3ex]
	\subfloat[\label{fig:4c}]{
		\begin{tabular}{c}
			\includegraphics[scale = 0.6]{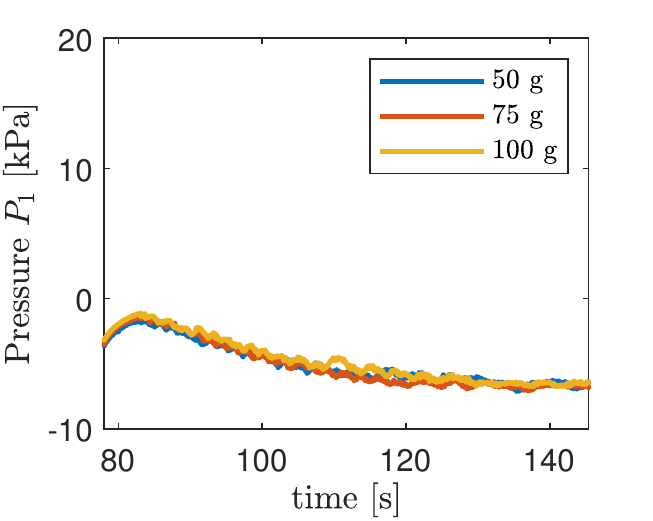} 
		\end{tabular}
	}
	\hspace*{-2.7em}
	\subfloat[\label{fig:4d}]{
		\begin{tabular}{c}
			\includegraphics[scale = 0.6]{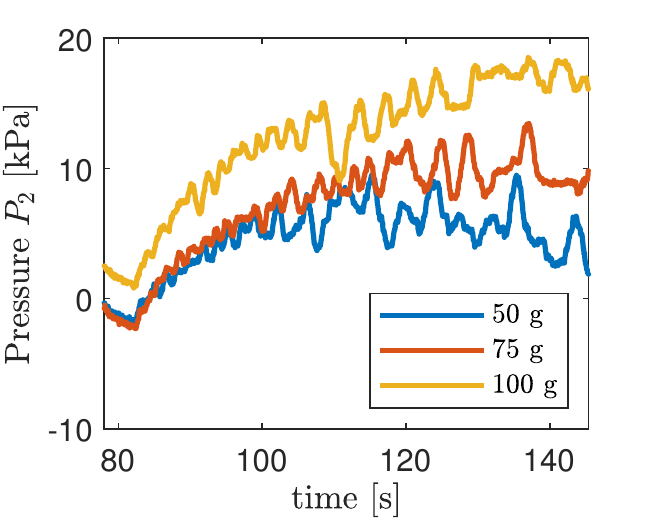} 
		\end{tabular}
	}\\[-0.3ex]
	\subfloat[\label{fig:4e}]{
		\begin{tabular}{c}
			\includegraphics[scale = 0.6]{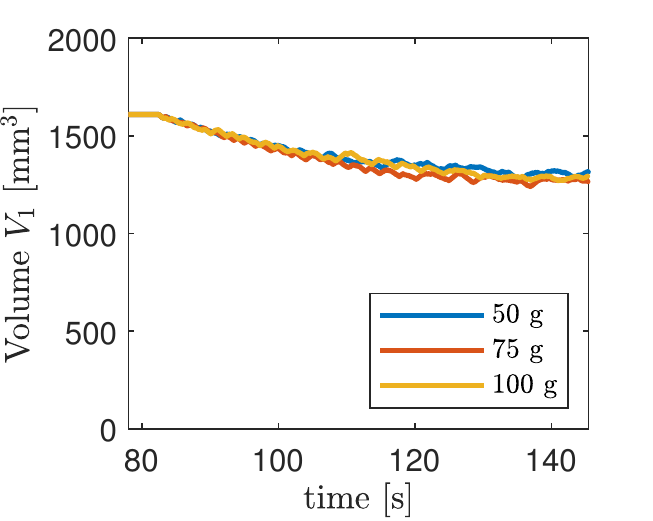} 
		\end{tabular}
	}
	\hspace*{-2.7em}
	\subfloat[\label{fig:4f}]{
		\begin{tabular}{c}
			\includegraphics[scale = 0.6]{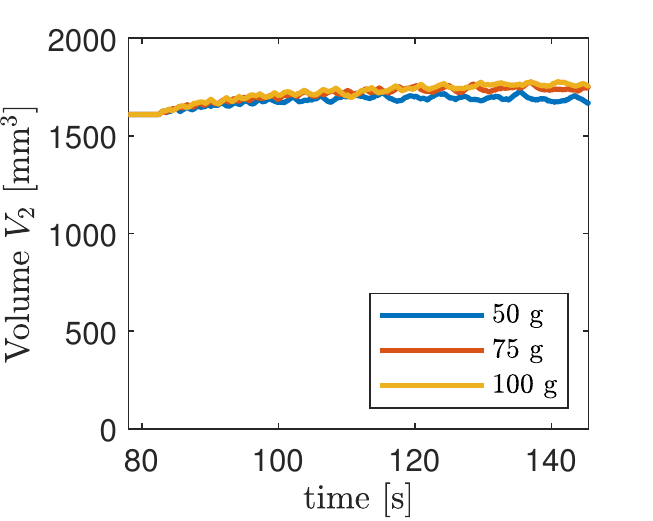} 
		\end{tabular}
	}
		\caption{Experimental results for system (\ref{eq:7}) with controller (\ref{eq:16}) under various loads:  (a) Position \(x\). (b) Force estimate \(\Tilde{F}\). (c) Pressure $P_{1}$. (d) Pressure $P_{2}$. (e) Stepper position $x_{s1}$. (f) Stepper position $x_{s2}$.} \label{fig:4} \vspace{-0.70cm}
	\end{center}
\end{figure}

Figure \ref{fig:4} shows that the controller achieves the regulation goal \(x=x^*\) with a consistent transient, which is comparable to the simulations in Figure \ref{fig:2}, regardless of the external forces.
Nevertheless, the settling time is larger than in the simulations since the syringe pumps and the stepper motors have not been accounted for in the controller design. The force estimate \(\Tilde{F}\) computed from (\ref{eq:9}) shows an initial spike, corresponding to the instant when the additional mass starts pulling on the actuators, and then settles around a constant value. Differently from Figure \ref{fig:2}, the disturbance estimate and the pressures are affected by high frequency noise which is due to: i) measurement noise on the position and the velocity (i.e., computed by discrete differentiation), which could be reduced with a low-pass filter; ii) quantization effects due to the stepper resolution, which could be improved by increasing the degree of microstepping used.

\begin{figure} [t!]
	\begin{center}
	\subfloat[\label{fig:5a}]{
		\begin{tabular}{c}
			\includegraphics[scale = 0.6]{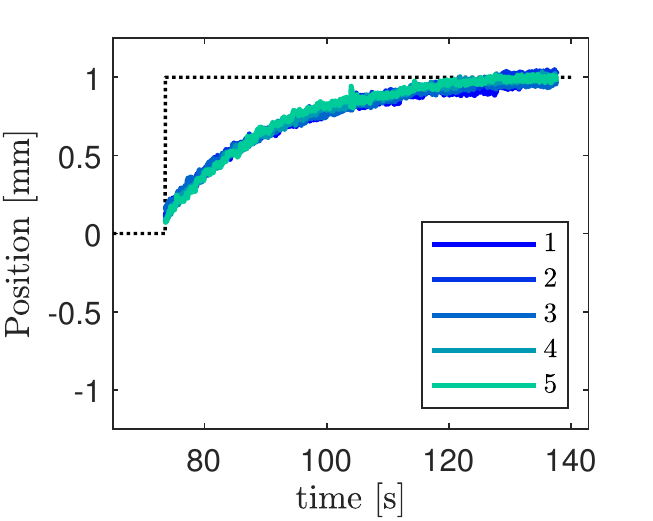} 
		\end{tabular}
	}
	\hspace*{-2.7em}
	\subfloat[\label{fig:5b}]{
		\begin{tabular}{c}
			\includegraphics[scale = 0.6]{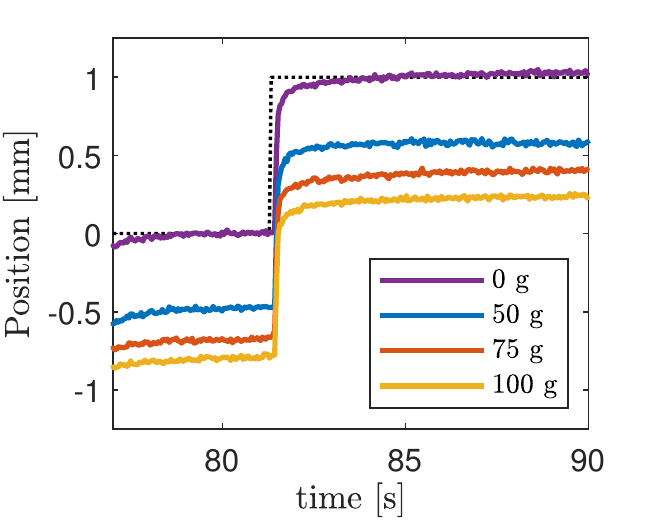} 
		\end{tabular}
	}
		\caption{Positioning results of (a) five repetitions of energy shaping controller (\ref{eq:16}) with 50 g mass; (b) open-loop control method \cite{Runciman2021} under various loads.} \label{fig:5} \vspace{-0.5cm}
	\end{center}
\end{figure}

As shown in Figure \ref{fig:5a}, the test with 50g mass was repeated five times demonstrating good repeatability, with a maximum standard deviation of the error with respect to the mean of 0.022 mm across any of the five repetitions.
The maximum error after settling (corresponding to t = 135 s in Figure \ref{fig:5a}) was 0.043 mm, and the mean error was 0.006 mm with a standard deviation of 0.033 mm.

Figure \ref{fig:5b} shows the system response with our open-loop controller \cite{Runciman2021} (i.e., the control input is related to the prescribed position \(x^*\) with a lookup table) in the presence of different payloads. In this case the external forces result in noticeable position errors, thus confirming that a feedback controller is necessary for high accuracy. 
Conversely, the open-loop approach \cite{Runciman2021}  yields a faster response: the reference position \(x_{sj}^*\) issued at the start of the experiments corresponds to the setpoint \(x^*\), thus the responsiveness of the system depends only on the maximum speed of the stepper motor. In addition, open-loop control is not affected by measurement noise from the tracking system. 

\begin{figure} [t!]
	\begin{center}
	\begin{tabular}{c}
		\includegraphics[scale = 0.6]{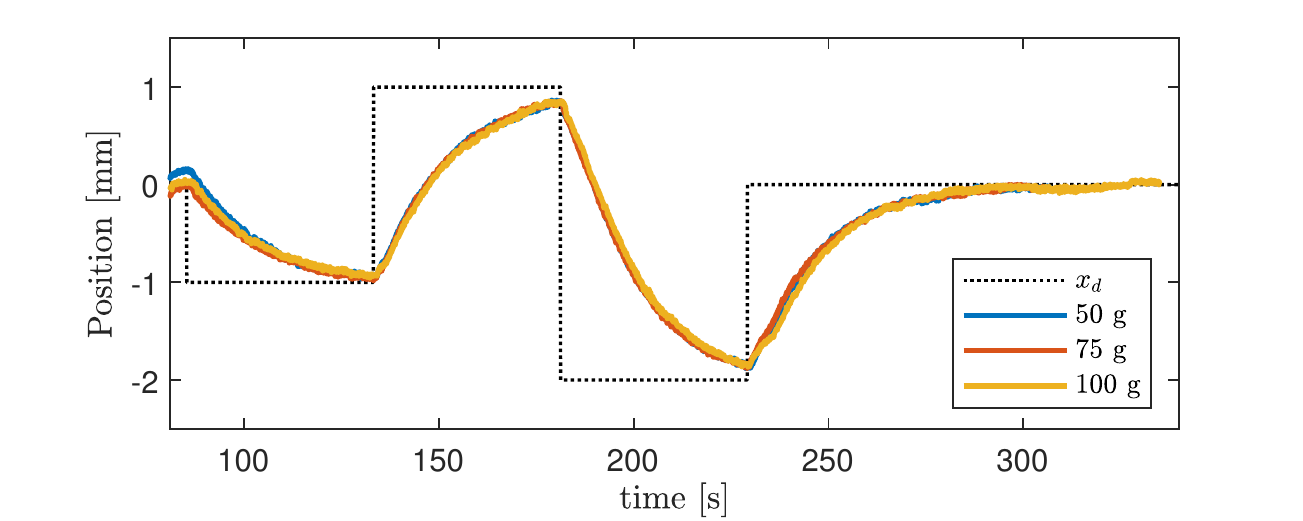} 
	\end{tabular}
	\caption{Experimental results of energy shaping controller (\ref{eq:16}) for multiple setpoints under various payloads.} \label{fig:6} \vspace{-0.5cm}
	\end{center}
\end{figure}

Figure \ref{fig:6} shows an additional set of results, where the same tuning parameters have been employed, but the setpoint \(x^*\) varies in time on both sides of the initial position. The results indicate that the proposed controller is effective in different operating conditions and yields a similar transient to Figure \(\ref{fig:4}\), suggesting that tuning does not need to be altered depending on \(x^*\), which is an important advantage in engineering practice. 
A video of the experiments has been included as a supplementary file.

\section{Conclusion}\label{sec:6}
In this work we have investigated the model based position control of a system consisting of two soft hydraulic bellow actuators arranged in an antagonistic pair. A dynamical model of the system, which includes the pressure dynamics of the fluid, has been defined in port-Hamiltonian form. A nonlinear observer has been designed to compensate the effect of external forces. A nonlinear control algorithm has then been constructed with an energy shaping approach. Although the control laws are specific to the antagonistic pair, the proposed approach can be readily extended to systems of multiple actuators arranged in different configurations. Therefore, there is potential to use this energy shaping control method to deliver force estimation and high accuracy positioning capabilities to rapidly manufactured, low-cost soft robotic systems, enabling many exciting applications.

The simulations results indicate that the controller achieves the prescribed regulation goal in the presence of different external forces. The experimental results on a prototype setup confirm that the controller can compensate the effect of model uncertainties and external forces, which have a similar magnitude to some MIS tasks, and that it yields higher accuracy compared to our previous open-loop approach.
In addition, the system response remains consistent across a range of operating conditions without needing to vary the tuning parameters, which is an advantage in engineering practice.
Conversely, the open-loop controller resulted in faster response, which depends only on the maximum speed of the stepper, and proved to be immune to sensor noise. However, these advantages come at the cost of having to individually characterize each actuator, which is time consuming and difficult to scale up. 
As such, future work will investigate a hybrid control approach with the aim of combining the advantages of both methods. 
In addition, we shall investigate 
more complex actuator arrangements.
\addtolength{\textheight}{-11cm}   




\bibliographystyle{IEEEtran}
\bibliography{reference.bib}
 
\end{document}

%% file: Figures/Figure1a.pdf_tex
\begingroup%
  \makeatletter%
  \providecommand\color[2][]{%
    \errmessage{(Inkscape) Color is used for the text in Inkscape, but the package 'color.sty' is not loaded}%
    \renewcommand\color[2][]{}%
  }%
  \providecommand\transparent[1]{%
    \errmessage{(Inkscape) Transparency is used (non-zero) for the text in Inkscape, but the package 'transparent.sty' is not loaded}%
    \renewcommand\transparent[1]{}%
  }%
  \providecommand\rotatebox[2]{#2}%
  \newcommand*\fsize{\dimexpr\f@size pt\relax}%
  \newcommand*\lineheight[1]{\fontsize{\fsize}{#1\fsize}\selectfont}%
  \ifx\svgwidth\undefined%
    \setlength{\unitlength}{200.55029441bp}%
    \ifx\svgscale\undefined%
      \relax%
    \else%
      \setlength{\unitlength}{\unitlength * \real{\svgscale}}%
    \fi%
  \else%
    \setlength{\unitlength}{\svgwidth}%
  \fi%
  \global\let\svgwidth\undefined%
  \global\let\svgscale\undefined%
  \makeatother%
  \begin{picture}(1,0.41686819)%
    \lineheight{1}%
    \setlength\tabcolsep{0pt}%
    \put(0,0){\includegraphics[width=\unitlength,page=1]{Figure1a.pdf}}%
    \put(0.77719601,0.21921091){\color[rgb]{0,0,0}\makebox(0,0)[lt]{\lineheight{1.25}\smash{\begin{tabular}[t]{l}$\theta$\end{tabular}}}}%
    \put(0.31591241,0.19748408){\color[rgb]{0,0,0}\makebox(0,0)[lt]{\lineheight{1.25}\smash{\begin{tabular}[t]{l}$L_{0}$\end{tabular}}}}%
    \put(0.95733018,0.19718712){\color[rgb]{0,0,0}\makebox(0,0)[lt]{\lineheight{1.25}\smash{\begin{tabular}[t]{l}$\frac{L_0}{3}$\end{tabular}}}}%
    \put(0.43863185,0.37761621){\color[rgb]{0,0,0}\makebox(0,0)[lt]{\lineheight{1.25}\smash{\begin{tabular}[t]{l}$d_{c}$\end{tabular}}}}%
    \put(0.64259489,0.37761621){\color[rgb]{0,0,0}\makebox(0,0)[lt]{\lineheight{1.25}\smash{\begin{tabular}[t]{l}$d_{c}$\end{tabular}}}}%
    \put(0.52899488,0.37761621){\color[rgb]{0,0,0}\makebox(0,0)[lt]{\lineheight{1.25}\smash{\begin{tabular}[t]{l}$D_{s}$\end{tabular}}}}%
  \end{picture}%
\endgroup%

%% file: Figures/Figure1b.pdf_tex
\begingroup%
  \makeatletter%
  \providecommand\color[2][]{%
    \errmessage{(Inkscape) Color is used for the text in Inkscape, but the package 'color.sty' is not loaded}%
    \renewcommand\color[2][]{}%
  }%
  \providecommand\transparent[1]{%
    \errmessage{(Inkscape) Transparency is used (non-zero) for the text in Inkscape, but the package 'transparent.sty' is not loaded}%
    \renewcommand\transparent[1]{}%
  }%
  \providecommand\rotatebox[2]{#2}%
  \newcommand*\fsize{\dimexpr\f@size pt\relax}%
  \newcommand*\lineheight[1]{\fontsize{\fsize}{#1\fsize}\selectfont}%
  \ifx\svgwidth\undefined%
    \setlength{\unitlength}{197.82066766bp}%
    \ifx\svgscale\undefined%
      \relax%
    \else%
      \setlength{\unitlength}{\unitlength * \real{\svgscale}}%
    \fi%
  \else%
    \setlength{\unitlength}{\svgwidth}%
  \fi%
  \global\let\svgwidth\undefined%
  \global\let\svgscale\undefined%
  \makeatother%
  \begin{picture}(1,0.51001417)%
    \lineheight{1}%
    \setlength\tabcolsep{0pt}%
    \put(0,0){\includegraphics[width=\unitlength,page=1]{Figure1b.pdf}}%
    \put(0.88014257,0.35547449){\color[rgb]{0,0,0}\makebox(0,0)[lt]{\lineheight{1.25}\smash{\begin{tabular}[t]{l}$V_{2}$\end{tabular}}}}%
    \put(0.23338005,0.21891867){\color[rgb]{1,0,0}\makebox(0,0)[lt]{\lineheight{1.25}\smash{\begin{tabular}[t]{l}$U_{1}$\end{tabular}}}}%
    \put(0.85002588,0.21891867){\color[rgb]{1,0,0}\makebox(0,0)[lt]{\lineheight{1.25}\smash{\begin{tabular}[t]{l}$U_{2}$\end{tabular}}}}%
    \put(0.54528487,0.46303464){\color[rgb]{0,0,0}\makebox(0,0)[lt]{\lineheight{1.25}\smash{\begin{tabular}[t]{l}$\dot{x}$\end{tabular}}}}%
    \put(0.48861803,0.26682266){\color[rgb]{0,0,0}\makebox(0,0)[lt]{\lineheight{1.25}\smash{\begin{tabular}[t]{l}$x$\end{tabular}}}}%
    \put(0.53275449,0.36311321){\color[rgb]{0,0,0}\makebox(0,0)[lt]{\lineheight{1.25}\smash{\begin{tabular}[t]{l}$m$\end{tabular}}}}%
    \put(0.55143202,0.03897767){\color[rgb]{0,0,0}\makebox(0,0)[lt]{\lineheight{1.25}\smash{\begin{tabular}[t]{l}Syringe pump 2 \end{tabular}}}}%
    \put(0.30577037,0.35547449){\color[rgb]{0,0,0}\makebox(0,0)[lt]{\lineheight{1.25}\smash{\begin{tabular}[t]{l}$V_{1}$\end{tabular}}}}%
    \put(0,0){\includegraphics[width=\unitlength,page=2]{Figure1b.pdf}}%
    \put(0.27566575,0.03969746){\color[rgb]{0,0,0}\makebox(0,0)[lt]{\lineheight{1.25}\smash{\begin{tabular}[t]{l}Syringe pump 1 \end{tabular}}}}%
    \put(0,0){\includegraphics[width=\unitlength,page=3]{Figure1b.pdf}}%
  \end{picture}%
\endgroup%

%% file: Figures/Setup.pdf_tex
\begingroup%
  \makeatletter%
  \providecommand\color[2][]{%
    \errmessage{(Inkscape) Color is used for the text in Inkscape, but the package 'color.sty' is not loaded}%
    \renewcommand\color[2][]{}%
  }%
  \providecommand\transparent[1]{%
    \errmessage{(Inkscape) Transparency is used (non-zero) for the text in Inkscape, but the package 'transparent.sty' is not loaded}%
    \renewcommand\transparent[1]{}%
  }%
  \providecommand\rotatebox[2]{#2}%
  \newcommand*\fsize{\dimexpr\f@size pt\relax}%
  \newcommand*\lineheight[1]{\fontsize{\fsize}{#1\fsize}\selectfont}%
  \ifx\svgwidth\undefined%
    \setlength{\unitlength}{295.61471053bp}%
    \ifx\svgscale\undefined%
      \relax%
    \else%
      \setlength{\unitlength}{\unitlength * \real{\svgscale}}%
    \fi%
  \else%
    \setlength{\unitlength}{\svgwidth}%
  \fi%
  \global\let\svgwidth\undefined%
  \global\let\svgscale\undefined%
  \makeatother%
  \begin{picture}(1,0.53911544)%
    \lineheight{1}%
    \setlength\tabcolsep{0pt}%
    \put(0,0){\includegraphics[width=\unitlength,page=1]{Setup.pdf}}%
    \put(0.09205163,0.00712243){\color[rgb]{0,0,0}\makebox(0,0)[lt]{\lineheight{1.25}\smash{\begin{tabular}[t]{l}$x$, $\dot{x}$\end{tabular}}}}%
    \put(0.06786668,0.50754609){\color[rgb]{0,0,0}\makebox(0,0)[lt]{\lineheight{1.25}\smash{\begin{tabular}[t]{l}Optical markers\end{tabular}}}}%
    \put(0.53426963,0.00748574){\color[rgb]{0,0,0}\makebox(0,0)[lt]{\lineheight{1.25}\smash{\begin{tabular}[t]{l}Gantry plate, $m$\end{tabular}}}}%
    \put(-0.00281624,0.28774399){\color[rgb]{0,0,0}\makebox(0,0)[lt]{\lineheight{1.25}\smash{\begin{tabular}[t]{l}$F_{ext}$\end{tabular}}}}%
    \put(0,0){\includegraphics[width=\unitlength,page=2]{Setup.pdf}}%
    \put(0.38387195,0.50974091){\color[rgb]{1,0,0}\makebox(0,0)[lt]{\lineheight{1.25}\smash{\begin{tabular}[t]{l}$U_{1}$\end{tabular}}}}%
    \put(0.68829683,0.51048069){\color[rgb]{1,0,0}\makebox(0,0)[lt]{\lineheight{1.25}\smash{\begin{tabular}[t]{l}$U_{2}$\end{tabular}}}}%
  \end{picture}%
\endgroup%